\documentclass[runningheads]{llncs}
\usepackage{graphicx}

\usepackage{tikz}
\usepackage{comment}
\usepackage{amsmath,amssymb} %
\usepackage{color}
\usepackage{subfig}
\usepackage{siunitx}

\newsavebox\CBox
\def\textBF#1{\sbox\CBox{#1}\resizebox{\wd\CBox}{\ht\CBox}{\textbf{#1}}}

\usepackage[hang,flushmargin]{footmisc}

\newcommand{\NEW}[1]{{\color{black}{#1}}}%
\newcommand{\mparagraph}[1]{\vspace{0.2em}\noindent\textbf{#1.}\hspace{1mm}}

\usepackage{ifpdf}
\usepackage{graphicx}

\ifpdf
  
\else
\fi
\usepackage{caption}
\usepackage{color}
\definecolor{cyan}{cmyk}{1,0,0,0}
\definecolor{darkgreen}{rgb}{0,0.5,0}
\definecolor{orange}{rgb}{1,0.5,0}
\definecolor{magenta}{cmyk}{0,1,0,0}
\definecolor{darkyellow}{cmyk}{0,0,0.75,0}
\definecolor{gray}{rgb}{0.8,0.8,0.8}

\usepackage{amsmath} %

\usepackage[toc,page]{appendix} %
\usepackage{wrapfig} %
\usepackage{comment}
\usepackage{cuted} %
\usepackage{lipsum} %
\usepackage{mathtools} %
\usepackage{gensymb}
\usepackage{float}
\usepackage{soul}
\usepackage{array}
\usepackage{multirow}
\usepackage{hhline}
\usepackage{setspace}
\usepackage{adjustbox}
\usepackage{booktabs}
\usepackage{orcidlink}

\usepackage[accsupp]{axessibility}  %

\begin{document}

\pagestyle{headings}
\mainmatter
\def\ECCVSubNumber{3503}  %

\graphicspath{{figs/}}
\title{FloatingFusion: Depth from ToF and Image-stabilized Stereo Cameras}

\author{Andreas Meuleman\inst{1}\orcidlink{0000-0002-9899-6365} \and
Hakyeong Kim\inst{1}\orcidlink{0000-0003-1061-7576} \and \\
James Tompkin\inst{2}\orcidlink{0000-0003-2218-2899}
\and
Min H. Kim\inst{1}\orcidlink{0000-0002-5078-4005}
}
\authorrunning{A. Meuleman et al.}
\institute{KAIST, South Korea\\
\email{\{ameuleman,hkkim,minhkim\}@vclab.kaist.ac.kr}\\
 \and
Brown University, United States\\
}

\maketitle

\begin{abstract}
High-accuracy per-pixel depth is vital for computational photography, so smartphones now have multimodal camera systems with time-of-flight (ToF) depth sensors and multiple color cameras.
However, producing accurate high-resolution depth is still challenging due to the low resolution and limited active illumination power of ToF sensors.
Fusing RGB stereo and ToF information is a promising direction to overcome these issues, but a key problem remains: to provide high-quality 2D RGB images, the main color sensor's lens is optically stabilized, resulting in an unknown pose for the floating lens that breaks the geometric relationships between the multimodal image sensors.
Leveraging ToF depth estimates and a wide-angle RGB camera, we design an automatic calibration technique based on dense 2D/3D matching that can estimate camera extrinsic, intrinsic, and distortion parameters of a stabilized main RGB sensor from a single snapshot. 
This lets us fuse stereo and ToF cues via a correlation volume.
For fusion, we apply deep learning via a real-world training dataset with depth supervision estimated by a neural reconstruction method. 
For evaluation, we acquire a test dataset using a commercial high-power depth camera and show that our approach achieves higher accuracy than existing baselines.

\keywords{Online camera calibration, 3D imaging, depth estimation, multi-modal sensor fusion, stereo imaging, time of flight.}
\end{abstract}

\section{Introduction}
\label{sec:intro}

Advances in computational photography allow many applications such as 3D reconstruction~\cite{Ha2021NormalFusion}, view synthesis~\cite{kopf2020oneshot3Dphoto,shih2020inpainting3Dphoto}, depth-aware image editing~\cite{Wadhwa2018,zhang2019defocus}, and augmented reality~\cite{Holynski2018densification,Valentin2018filter}. 
Vital to these algorithms is \emph{high-accuracy per-pixel depth}, e.g., to integrate virtual objects by backprojecting high-resolution camera color into 3D.
To this end, smartphones now have camera systems with multiple sensors, lenses of different focal lengths, and active-illumination time-of-flight (ToF). 
For instance, correlation-based ToF provides depth by measuring the travel time of infrared active illumination with a gated infrared sensor.

\begin{figure*}[t]
	\centering
	\resizebox{\textwidth}{!}{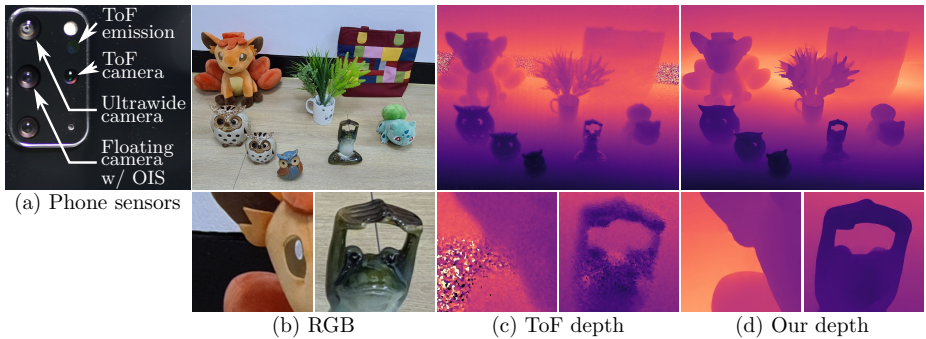}
	\vspace{-0.5cm}
	\caption{\label{fig:teaser}
		(a) Multi-modal smartphone imaging. (b) Reference RGB image. (c) ToF depth reprojected to the reference. (d) Our depth from floating fusion.
	}
	\vspace{-0.25cm} %
\end{figure*}

We consider two challenges in providing high-accuracy per-pixel depth: 
\NEW{(1) ToF} sensor spatial resolution is orders of magnitude less than that of its compatriot color cameras. RGB spatial resolution has increased dramatically on smartphones---12--64 million pixels is common---whereas ToF is often 0.05--0.3 million pixels. 
One might correctly think that fusing depth information from ToF with depth information from color camera stereo disparity is a good strategy to increase our depth resolution. Fusion might also help us overcome the low signal-to-noise ratio in ToF signals that arises from the low-intensity illumination of a battery-powered device. For fusion, we need to accurately know the geometric poses of all sensors and lenses in the camera system.

This leads to our second challenge: 
(2)~As RGB spatial resolution has increased, smartphones now use optical image stabilization \cite{sachs2006image,Wang2017phoneIOS}: a \emph{floating} lens compensates for camera body motion to avoid motion blur during exposure.
Two low-power actuators suspend the lens body vertically and horizontally to provide a few degrees of in-plane rotation or translation, similar to how a third actuator translates the lens along the optical axis for focus. 
The magnetic actuation varies with focus and even with the smartphone's orientation due to gravity, and the pose of the stabilizer is not currently possible to measure or read out \NEW{electronically}. As such, we can only use a fusion strategy if we can automatically optically calibrate the geometry of the floating lens for each exposure taken.

This work proposes a \emph{floating fusion} algorithm to provide high accuracy per pixel depth estimates from an optically-image-stabilized camera, a second RGB camera, and a ToF camera (Fig.~\ref{fig:teaser}).
We design an online calibration approach for the floating lens that uses ToF measurements and dense optical flow matching between the RGB camera pair. 
This lets us form 2D/3D correspondences to recover intrinsic, extrinsic, and lens distortion parameters in an absolute manner (not `up to scale'), and for every snapshot. This makes it suitable for dynamic environments.
Then, to fuse multi-modal sensor information, we build a correlation volume that integrates both ToF and stereo RGB cues, \NEW{then predict disparity via a learned function. There are few large multi-modal datasets to train this function, and synthetic data creation is expensive and retains a domain gap to the real world.}
Instead, we capture real-world scenes with multiple views and optimize a neural radiance field~\cite{barron2021mipnerf} with ToF supervision. The resulting depth maps are lower noise and higher detail than those of a depth camera\NEW{, and provide us with high-quality training data.}
For validation, we build a test dataset using a Kinect Azure and show that our method outperforms other traditional and data-driven approaches for snapshot RGB-D imaging.

\section{Related Work}

\mparagraph{ToF and RGB Fusion}
Existing data-driven approaches~\cite{agresti2017fusion,agresti2019fusion_synthetic,pham2019deep_fusion} heavily rely on synthetic data, creating a domain gap. This is exacerbated when using imperfect low-power sensors such on mobile phones.
In addition, current stereo-ToF fusion~\cite{evangelidis2015fusion,mutto2015probabilisticFusion,gao2017selfcalibratingtof} typically estimates disparity from stereo and ToF separately before fusion. One approach is to estimate stereo and ToF confidence to merge the disparity maps \cite{Marin2016ReliableFO,agresti2017fusion,agresti2019fusion_synthetic,poggi2020fusion_conf}.
In contrast, our ToF estimates are directly incorporated into our disparity pipeline before depth selection. 
\NEW{Fusion without stereo~\cite{jung2021wildtofu} tackles more challenging scenarios than direct ToF depth estimation. However, Jung et al.'s downsampling process can blur over occlusion edges, producing incorrect depth at a low resolution that is difficult to fix after reprojection at finer resolutions.}

\mparagraph{Phone and Multi-Sensor Calibration}
DiVerdi and Barron~\cite{DiVerdi2016calib} tackle per shot stereo calibration up to scale in the challenging mobile camera environment; however, absolute calibration is critical for stereo/ToF fusion. We leverage coarse ToF depth estimates for absolute stereo calibration.
\NEW{Gil et al.~\cite{gil2021OnlineCalibration} estimate two-view stereo calibration by first estimating a monocular depth map in one image before optimizing the differentiable projective transformation (DPT) that maximizes the consistency between the stereo depth and the monocular depth. The method refines the DPT parameters, handling camera pose shift after factory calibration and improving stereo depth quality, but it still requires the initial transformation to be sufficiently accurate for reasonable stereo depth estimation. In addition, to allow for stable optimization, a lower degree of freedom model is selected, which can neglect camera distortion and lens shift.}
Works tackling calibration with phone and ToF sensors are not common.
Gao et al.~\cite{gao2017selfcalibratingtof} use Kinect RGB-D inputs, match RGB to the other camera, use depth to lift points to 3D, then solves a PnP problem to find the transformation.
Since it matches sparse keypoints, it is not guaranteed that depth is available where a keypoint is, leading to too few available keypoints. In addition, the method does not account for intrinsic or distortion refinement.

\mparagraph{Data-Driven ToF Depth Estimation}
Numerous works~\cite{son2016tof,marco2017deeptof,guo2018tof_flat,agresti2018tof,su2018deeptof,gao2021tof} attempt to tackle ToF depth estimation via learned approaches. 
While these approaches have demonstrated strong capabilities in handling challenging artifacts (noise, multi-path interference, or motion), our approach does not strictly require a dedicated method for ToF depth estimation as we directly merge ToF samples in our stereo fusion pipeline.

\mparagraph{Conventional Datasets}
Accurate real-world datasets with ground-truth depth maps are common for stereo depth estimation~\cite{middlebury2014dataset,Menze2015CVPR,sunrgbd2015dataset}. However, the variety of fusion systems makes it challenging to acquire large-high-quality, real-world datasets. 
A majority of ToF-related works leverage rendered data~\cite{marco2017deeptof,guo2018tof_flat}, particularly for fusion datasets~\cite{agresti2017fusion,agresti2019fusion_synthetic}. These datasets enable improvement over conventional approaches, but synthesizing RGB and ToF images accurately is challenging. A domain gap is introduced as the noise profile and imaging artifacts are different from the training data. 
Notable exceptions are Son et al.~\cite{son2016tof}, and Gao and Fan et al.~\cite{gao2021tof}, where an accurate depth camera provides training data for a lower-quality ToF module. The acquisition is partially automated thanks to a robotic arm. However, this bulky setup limits the variety of the scenes: all scenes are captured on the same table, with similar backgrounds across the dataset. In addition, the use of a single depth camera at a different location from the ToF module introduces occlusion, with some areas in the ToF image having no supervision. In addition, this method only tackles ToF depth estimation, and the dataset does not feature RGB images.

\mparagraph{Multiview Geometry Estimation}
Several approaches are capable of accurate depth estimation from multiview images~\cite{schonberger2016colmap}, even in dynamic environments~\cite{li2019mannequin,luo2020cvd,kopf2021rcvd}. Despite their accuracy, including ToF data to these approaches is not obvious.
Scene representations optimized from a set of images~\cite{barron2021mipnerf,jeong2021SCNeRF,yu2021plenoxels} have recently shown good novel view synthesis and scene geometry reconstruction, including to refine depth estimates in the context of multiview stereo~\cite{wei2021nerfingmvs}. 
Since the optimization can accept supervision from varied sources, including ToF measurements\NEW{ is straightforward.} %
For this reason, we select a state-of-the-art neural representation that has the advantage to handle heterogeneous resolutions~\cite{barron2021mipnerf} for our training data generation.
\NEW{TöRF~\cite{attal2021torf} renders phasor images from a volume representation to optimize raw ToF image reconstruction. While efficiently improving NeRF's results and tackling ToF phase wrapping, this approach is not necessary for our context as our device is not prone to phase wrapping due to its low illumination range (low power) and thanks to the use of several modulation frequencies. We also observe that, in the absence of explicit ToF confidence, erroneous ToF measurements tend to be more present in depth maps rendered from a TöRF.
}
Finally, approaches based on ICP registration~\cite{Ha2021NormalFusion} cannot be applied directly to our data since depth maps from the low-power ToF module are too noisy to be registered through ICP.

\begin{figure*}[tp]
    \centering
    \resizebox{\textwidth}{!}{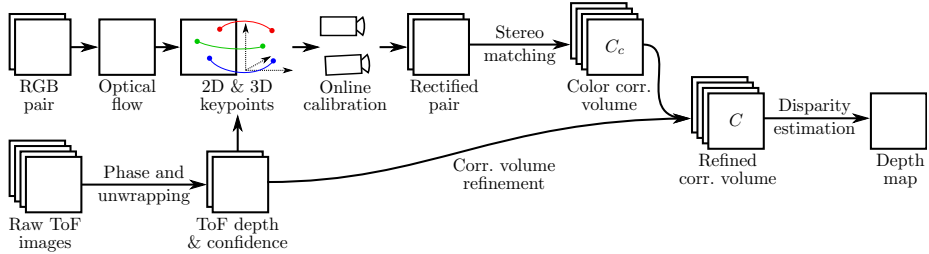}
    \caption{\label{fig:overview}
        Overview of our method. The first step is to estimate ToF depth via phase unwrapping ($\S$~\ref{sec:tof_depth}). Then, and after dense matching between the two RGB cameras, we use ToF depth to estimate the floating camera's intrinsic and extrinsic parameters ($\S$~\ref{sec:calib}). Once the stereo pair is calibrated and rectified, we extract features in each image to build a correlation volume $C_c$. This volume is refined using the ToF samples ($\S$~\ref{sec:fusion}) before the final disparity estimation.
    }
\end{figure*}

\section{Method}

We use an off-the-shelf Samsung Galaxy S20+ smartphone. This has the main camera with a 12MP color sensor and a magnetic mount 79° lens for stabilization and focusing, a secondary 12MP color camera with a fixed ultrawide 120° lens, and a 0.3MP ToF system with an infrared fixed 78° lens camera and infrared emitter (Fig.~\ref{fig:teaser}a).
As the ultrawide camera and the ToF module are rigidly fixed, we calibrate their intrinsics $K_{\text{UW}},K_{\text{ToF}}$, extrinsics $[R|t]_{\text{UW}},[R|t]_{\text{ToF}}$, and lens distortion parameters using an offline method based on checkerboard corner estimation. We use a checkerboard with similar absorption in the visible spectrum as in infrared. 
However, calibrating the floating main camera (subscript $_{\text{FM}}$) is not possible offline, as its pose changes from snapshot to snapshot. OIS introduces lens shift in $x,y$ for stabilization and in $z$ for focus, with the $z$ direction inducing additional lens distortion changes. The lens also tilts (pitch/yaw rotations) depending on the phone's orientation because of gravity.
As such, we must estimate per snapshot a new intrinsic matrix $K_{\text{FM}}$, new extrinsic matrix $[R|t]_{\text{FM}}$, and three radial and two tangential distortion coefficients $\{k_1,k_2,k_3,p_1,p_2\}_\text{FM}$ from the Brown-Conrady model~\cite{conrady1919lens,brown1966lens} for the main floating camera.
To tackle this challenge, we present a method to estimate these parameters at an absolute scale (not relative or `up to scale').

\subsection{ToF Depth Estimation}
\label{sec:tof_depth}

\begin{figure*}[t]
    \centering
    \resizebox{\textwidth}{!}{\begingroup%
  \makeatletter%
  \providecommand\color[2][]{%
    \errmessage{(Inkscape) Color is used for the text in Inkscape, but the package 'color.sty' is not loaded}%
    \renewcommand\color[2][]{}%
  }%
  \providecommand\transparent[1]{%
    \errmessage{(Inkscape) Transparency is used (non-zero) for the text in Inkscape, but the package 'transparent.sty' is not loaded}%
    \renewcommand\transparent[1]{}%
  }%
  \providecommand\rotatebox[2]{#2}%
  \newcommand*\fsize{\dimexpr\f@size pt\relax}%
  \newcommand*\lineheight[1]{\fontsize{\fsize}{#1\fsize}\selectfont}%
  \ifx\svgwidth\undefined%
    \setlength{\unitlength}{425.19685039bp}%
    \ifx\svgscale\undefined%
      \relax%
    \else%
      \setlength{\unitlength}{\unitlength * \real{\svgscale}}%
    \fi%
  \else%
    \setlength{\unitlength}{\svgwidth}%
  \fi%
  \global\let\svgwidth\undefined%
  \global\let\svgscale\undefined%
  \makeatother%
  \begin{picture}(1,0.29333333)%
    \lineheight{1}%
    \setlength\tabcolsep{0pt}%
    \put(0,0){\includegraphics[width=\unitlength,page=1]{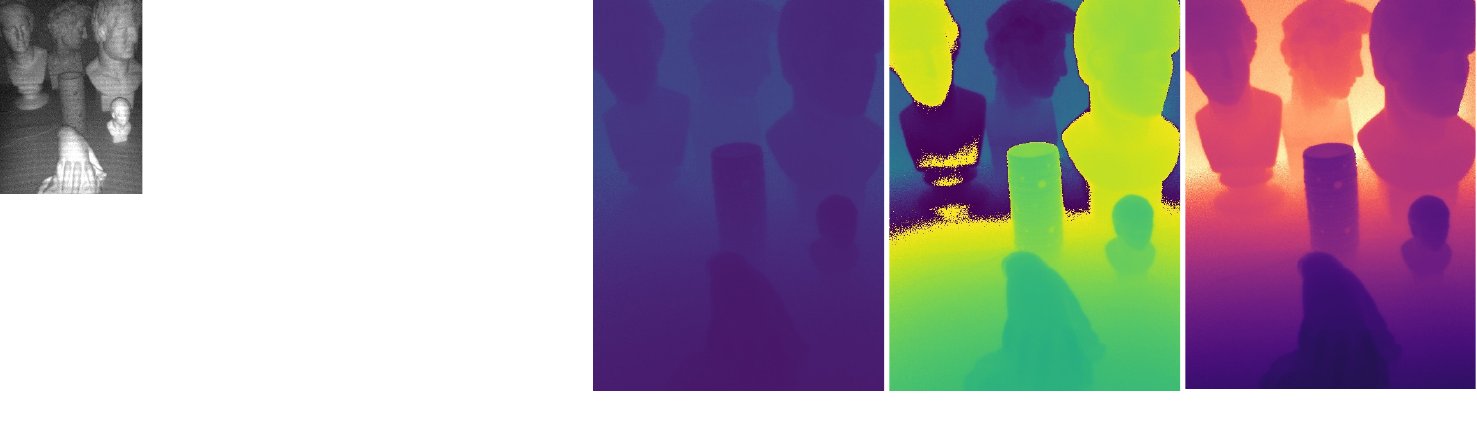}}%
    \put(0.2018973,0.00742472){\color[rgb]{0,0,0}\makebox(0,0)[t]{\smash{\begin{tabular}[t]{c}(a) Raw ToF measurements ($Q_{f,\theta}$) \end{tabular}}}}%
    \put(0.50029388,0.00742472){\color[rgb]{0,0,0}\makebox(0,0)[t]{\smash{\begin{tabular}[t]{c}(b) $\phi_{20\text{MHz}}$\end{tabular}}}}%
    \put(0.70081781,0.00742472){\color[rgb]{0,0,0}\makebox(0,0)[t]{\smash{\begin{tabular}[t]{c}(c) $\phi_{100\text{MHz}}$\end{tabular}}}}%
    \put(0.90134176,0.00742472){\color[rgb]{0,0,0}\makebox(0,0)[t]{\smash{\begin{tabular}[t]{c}(d) ToF depth\\\end{tabular}}}}%
    \put(0,0){\includegraphics[width=\unitlength,page=2]{tof_breakdown.pdf}}%
  \end{picture}%
\endgroup%
}
    \caption{\label{fig:tof}
       ToF depth estimation from raw measurements. From raw ToF images for two different frequencies (a), we estimate a coarse but unwrapped phase map~(b) and a finer but wrapped phase map (c). By unwrapping $\phi_{100\text{MHz}}$ using the lower frequency phase $\phi_{20\text{MHz}}$, we estimate a more accurate depth map.
    }
\end{figure*}

The ToF system modulates its infra-red light source by 20MHz and 100MHz square waves, and alternates between both frequencies sequentially. The sensor captures four shots per frequency, with the same modulation as the light source shifted by 90 degrees. With two frequencies, we obtain eight raw ToF measurements per snapshot:
\begin{equation}
    Q_{f,\theta},\ \hspace{4mm} f\in \{20\text{MHz}, 100\text{MHz}\}, \hspace{4mm}  \theta \in \{0, \pi/2, \pi, 3\pi/2\},
\end{equation}
where $f$ is the modulation frequency and $\theta$ is the sensor phase shift.
Figure~\ref{fig:tof}(a) shows an example of captured ToF measurements.
To estimate depth, the first step is to estimate phase (Figures~\ref{fig:tof}(b) and~\ref{fig:tof}(c)) for each modulation frequency:
\begin{equation}
    \phi_f = \text{arctan2}\left(Q_{f,\pi} - Q_{f,0}, Q_{f,3\frac{\pi}{2}} - Q_{f,\frac{\pi}{2}}\right).
\end{equation}
From these phases, we can estimate the distance:
\begin{equation}
    \label{eq:phase2depth}
    d_f = \frac{c}{4\pi f}\phi_f + k_f\frac{c}{2f}, \hspace{4mm} k_f\in\mathbb{N},
\end{equation}
with $c$ the speed of light.
This equation shows that, given a phase, the depth is known up to a $k_f\frac{c}{2f}$ shift. That is, we observe phase wrapping.
The ToF depth variance is inversely proportional to the modulation frequency \cite{Li2014tofTIintro}. Therefore, the higher frequency tends to be more accurate; however, the wrapping range is shorter ($\approx$1.5m for 100MHz against $\approx$7.5m for 20MHz).
Since the ToF illumination is low power, the signal becomes weak for further objects, making the phase estimation highly unreliable beyond the wrapping point of the lower frequency. Based on this, we assume that $\phi_{20\text{MHz}}$ does not show phase wrapping: $k_{20\text{MHz}} := 0$.
Therefore, we unwrap $\phi_{100\text{MHz}}$ using $\phi_{20\text{MHz}}$ to benefit from the lower depth variance associated with a higher frequency without the phase wrapping ambiguity. In detail, we find $\hat k_{100\text{MHz}}$ that minimizes the depth difference between the two frequencies:
\begin{equation}
    \hat k_{100\text{MHz}} = \underset{k}{\text{argmin}}\left|d_{20\text{MHz}} - \frac{c}{4\pi \cdot10^8}\phi_{100\text{MHz}} + k\frac{c}{2\cdot10^8} \right|.
\end{equation}
From this, we can compute the distance $d_{100\text{MHz}}$, which will be used to obtain the depth $d_{ToF}$ (Figure~\ref{fig:tof}d).
For more details on ToF, refer to~\cite{hansard2012tofbook,Li2014tofTIintro}.

With the estimated depth, we assign a confidence map $\omega$ based on the signal's amplitude, the concordance between $d_{20\text{MHz}}$ and $d_{100\text{MHz}}$ and local depth changes.
First, we assign lower scores when the signal is weak:
\begin{equation}
    \omega_A = \exp\left(-{1}/\sum_{f}A_f / (2 \sigma_A^2)\right),
\end{equation}
With $A_f = \sqrt{(Q_{f,0} - Q_{f,\frac{\pi}{2}})^2 + (Q_{f,\pi} - Q_{f,3\frac{\pi}{2}})^2} / 2$ and $f \in \{20\text{MHz}, 100\text{MHz}\}$. We also take into account the difference between the estimated distance from the two frequencies:
\begin{equation}
    \omega_d = \exp\left(-|d_{20\text{MHz}} - d_{100\text{MHz}}|^2 / (2 \sigma_d^2)\right).
\end{equation}
Since ToF is less reliable at depth discontinuities, we deem areas with large depth gradient to be less reliable:
\begin{equation}
    \omega_\nabla = \exp\left(-\|\nabla d_\text{ToF}\|^2 / (2 \sigma_\nabla^2)\right) \cdot \exp\left(-\|\nabla (1/d_\text{ToF})\|^2 / (2 \sigma_\nabla^2)\right).
\end{equation}
The complete confidence is: $\omega = \omega_A \omega_d \omega_\nabla$. For our experiments, we set $\sigma_A = 20$, $\sigma_d = 0.05$, and $\sigma_\nabla = 0.005$.

\subsection{Online Calibration}
\label{sec:calib}

To calibrate our floating main camera, we need to find sufficient correspondences between known 3D world points and projections of those points in 2D.
We must find a way to correspond our 3D points from ToF with the main camera even though it does not share a spectral response.
Our overall strategy is to use the known fixed relationship between the ToF and ultrawide cameras and additionally exploit 2D color correspondences from the optical flow between the ultrawide and floating cameras.
In this way, we can map from main camera 2D coordinates to ultrawide camera 2D coordinates to ToF camera 3D coordinates. While ToF can be noisy, it still provides sufficient points to robustly calibrate all intrinsic, extrinsic, and lens distortion parameters of the main camera.

\mparagraph{From ToF to Ultrawide}
The first step is to reproject the ToF depth estimates $d_{\text{ToF}}$ to the ultrawide camera. We transform the depth map to a point cloud $P$: 
\begin{equation}
    \label{eq:d_to_points}
    P_{\text{ToF}} = K^{-1}_{\text{ToF}} [u, v, d_{\text{ToF}}]^\top,
\end{equation}
where $K_{\text{ToF}}$ is the known ToF camera matrix, and $(u, v)$ are pixel coordinates with corresponding depth $d_{\text{ToF}}$. Then, the point cloud can be transformed to the ultrawide camera's space:
\begin{equation}
    \label{eq:space_transform}
    P_{\text{UW}} = [R|t]_{\text{ToF} \rightarrow \text{UW}} [P_{\text{ToF}}^T|1]^\top,
\end{equation}    
where $[R|t]_{ToF \rightarrow \text{UW}}$ is the relative transformation from the ToF camera space to the ultrawide camera space. From point cloud $P_{\text{UW}}$, we obtain the pixel coordinate $[u_{\text{UW}}, v_{\text{UW}}]$ of the ToF point cloud reprojected to the ultrawide camera: 
\begin{equation}
    \label{eq:points_2_uv}
    d_{\text{UW}} [u_{\text{UW}}, v_{\text{UW}}, 1] =  K_{\text{UW}} P_{\text{UW}}.   
\end{equation}
The reprojected ToF points $P_{\text{UW}}$ and their subpixel coordinates $[u_{\text{UW}}, v_{\text{UW}}]$ will be used to estimate calibration for our main camera in a later stage.

\mparagraph{From Ultrawide to Floating Main}
Next, we match the ultrawide camera to the floating main camera to be accurately calibrated. Since both cameras are located near to each other, they share a similar point of view, thus making sparse scale and rotation invariant feature matching unnecessary. As such, we use dense optical flow~\cite{teed2020raft} to find correspondences. To use flow, we first undistort the ultrawide camera image given its calibration, then rectify it \emph{approximately} to the floating main camera given an initial \emph{approximate} offline calibration. This calibration will be wrong, but as flow is designed for small unconstrained image-to-image correspondence, this rectification approach will still find useful correspondences.
As output, we receive a 2D vector field $\mathcal{F}_{\text{UW}\rightarrow \text{FM}}$.

\mparagraph{From ToF to Floating Main}
We can now use the optical flow to form 2D/3D matches.
For each ToF point reprojected to the ultrawide camera $P_{\text{UW}}$, we find the corresponding pixel in the floating camera: 
\begin{equation}
    \label{eq:uw_to_fm}
    [u'_{\text{FM}}, v'_{\text{FM}}] = [u_{\text{UW}}, v_{\text{UW}}] + \mathcal{F}_{\text{UW}\rightarrow \text{FM}}([u_{\text{UW}}, v_{\text{UW}}]). 
\end{equation}    
We sample the flow using bilinear interpolation since $[u_{\text{UW}}, v_{\text{UW}}]$ are estimated with subpixel precision.

From this 2D/3D matching between the 2D points in the floating camera $[u'_{\text{FM}}, v'_{\text{FM}}]$ \NEW{and }the 3D points in the ultrawide camera space $P_{\text{UW}}$, we can estimate the floating camera's calibration. We first solve the optimization through RANSAC for outlier removal, followed by Levenberg-Marquardt optimization,
 obtaining the transformation between the two RGB cameras $[R, t]_{\text{FM} \rightarrow \text{UW}}$, as well as the camera matrix $K_{\text{FM}}$ and its distortion coefficients. Once the calibration is achieved, we rectify the two RGB images, enabling stereo/ToF fusion.

\mparagraph{Discussion}
Both Equations~\eqref{eq:points_2_uv} and~\eqref{eq:uw_to_fm} are not occlusion-aware but, due to the small baseline, occluded world points are only a small portion of the total number of matched points. RANSAC helps us to avoid outlier correspondences from occlusion, incorrect flow estimates, and noisy ToF estimates for calibration.

Further, while we rely on a fixed RGB camera and ToF module, extending the approach to scenarios without fixed cameras is possible. Reliable feature matching between spectral domains has been demonstrated~\cite{efe2021dfm,zhu2019rgb_ir_matching,brown2011MultispectralSF} as well as RGB/IR optical flow~\cite{qiu2019tof_upsampling}. Using matching between the ToF module's IR camera and the RGB cameras, calibration can likely be achieved even if no second RGB camera exists as fixed with respect to the ToF module.

\subsection{Fusing ToF and Stereo}
\label{sec:fusion}

Given the now-calibrated color stereo pair, and the ToF depth samples, we will fuse these into an accurate high-resolution depth map for a color camera. The first step is to build a correlation volume $\mathcal{C}_c$ from our RGB pair. A point in the volume $\mathcal{C}_c$ at coordinate $[u, v, u']$ represents the correlation between a pixel $[u, v]$ in the reference image and a pixel $[u', v]$ in the target image at some disparity. Thus, the correlation volume's shape is (width$\times$height$\times$width) since disparity is horizontal along the width direction.
We compute correlation volume values by extracting 256-dim.~image features from each view using RAFT-stereo~\cite{lipson2021raft-stereo}'s learned feature encoder, then by taking the dot product of the feature vectors from each RGB camera for each disparity amount. 
\NEW{Note that the feature extraction process downsamples the images four times and a disparity map at original resolution is recovered through RAFT-stereo's convex upsampling.}
For ease of evaluation, we use the fixed ultrawide camera as a reference, although the floating camera can be chosen without algorithmic change.

A 3D world point $P_\text{ToF}$ corresponds to a 2D sensor coordinate $[u_{\text{UW}}, v_{\text{UW}}]$ in the ultrawide camera and in the floating main camera $[u_{\text{FM}}, v_{\text{FM}}]$ (see reprojection Equations~\eqref{eq:d_to_points},\eqref{eq:space_transform}, and \eqref{eq:points_2_uv})). A ToF point also has a confidence $\omega$ estimated along with the ToF depth maps. We leverage those points by increasing the correlation at the location of the sample points in the volume. For a point, the corresponding location in the volume is $[u_{\text{UW}}, v_{\text{UW}}, u_{\text{FM}}]$.
Coordinates given by a ToF point $[u_{\text{UW}}, v_{\text{UW}}, u_{\text{FM}}]$ have eight integer neighbors with defined values in the correlation volume. We inject the ToF point into RAFT-stereo's correlation volume using linear-like weights:
\begin{equation}
\small
\begin{split}
    \mathcal{C}(\lfloor u_{\text{UW}}\rfloor, \lfloor v_{\text{UW}}\rfloor, \lfloor u_{\text{FM}}\rfloor) = &\ \mathcal{C}_c(\lfloor u_{\text{UW}}\rfloor, \lfloor v_{\text{UW}}\rfloor, \lfloor u_{\text{FM}}\rfloor) ~+ \\ 
    & \tau \cdot \omega \left((\lceil u_{\text{UW}}\rceil - u_{\text{UW}})(\lceil v_{\text{UW}}\rceil - v_{\text{UW}})(\lceil u_{\text{FM}}\rceil - u_{\text{FM}}) \right).
\end{split}
\end{equation}
$\lfloor u\rfloor$ is the nearest integer $\leq$ u, and $\lceil \rceil$ is the nearest integer $\geq$ u. We perform the same operation for all eight integer points with coordinates around $[u, v, u']$. If several ToF points affect the same point in the correlation, their contributions accumulate. $\tau$ is a scalar parameter that is optimized during the training process. We then use the updated correlation volume $\mathcal{C}$ with the next steps of RAFT-stereo's pipeline to estimate disparity. Thanks to this approach, we efficiently and robustly combine stereo correlation cues and ToF measurements before estimating the depth map.

\subsection{Dataset Generation}
\label{sec:dataset}

\mparagraph{Background}
Learning-based stereo/ToF fusion methods require training data. For this, we optimize neural radiance fields~\cite{Lombardi2019neuralVolumes,mildenhall2020nerf,Lombardi2021neuralmixture} from RGB images. These allow querying density and appearance for every point in the scene, allowing us to render estimated depth maps. A typical volume rendering is as follows:
\begin{equation}
    \label{eq:rendering}
    \hat{C}(r) = \sum_{i=1}^N T_i(1-\exp(-\sigma_i\delta_i))c_i,
\end{equation}
\NEW{where $\hat{C}(r)$ is the rendered color for the ray $r$,
$\sigma_i$ and $c_i$ are the density and color of the representation at $i$ point. $N$ points are sampled along the ray $r$, where $\delta_i$ is the distance between neighboring samples, and $T_i=\exp\left(-\sum\nolimits_{i'<i}\sigma_{i'}\delta_{i'}\right)$ is the approximated transparancy between the ray origin and the sample.}

This rendering is differentiable on the density and color of the sample points, allowing gradient-based optimization. 
To render depth maps from these representations, we swap the color term $c_i$ in Equation (\ref{eq:rendering}) for the depth of the point w.r.t.~the ray origin.

\begin{figure}[t]
    \centering
    \resizebox{\columnwidth}{!}{\begingroup%
  \makeatletter%
  \providecommand\color[2][]{%
    \errmessage{(Inkscape) Color is used for the text in Inkscape, but the package 'color.sty' is not loaded}%
    \renewcommand\color[2][]{}%
  }%
  \providecommand\transparent[1]{%
    \errmessage{(Inkscape) Transparency is used (non-zero) for the text in Inkscape, but the package 'transparent.sty' is not loaded}%
    \renewcommand\transparent[1]{}%
  }%
  \providecommand\rotatebox[2]{#2}%
  \newcommand*\fsize{\dimexpr\f@size pt\relax}%
  \newcommand*\lineheight[1]{\fontsize{\fsize}{#1\fsize}\selectfont}%
  \ifx\svgwidth\undefined%
    \setlength{\unitlength}{425.19685039bp}%
    \ifx\svgscale\undefined%
      \relax%
    \else%
      \setlength{\unitlength}{\unitlength * \real{\svgscale}}%
    \fi%
  \else%
    \setlength{\unitlength}{\svgwidth}%
  \fi%
  \global\let\svgwidth\undefined%
  \global\let\svgscale\undefined%
  \makeatother%
  \begin{picture}(1,0.30322001)%
    \lineheight{1}%
    \setlength\tabcolsep{0pt}%
    \put(0,0){\includegraphics[width=\unitlength,page=1]{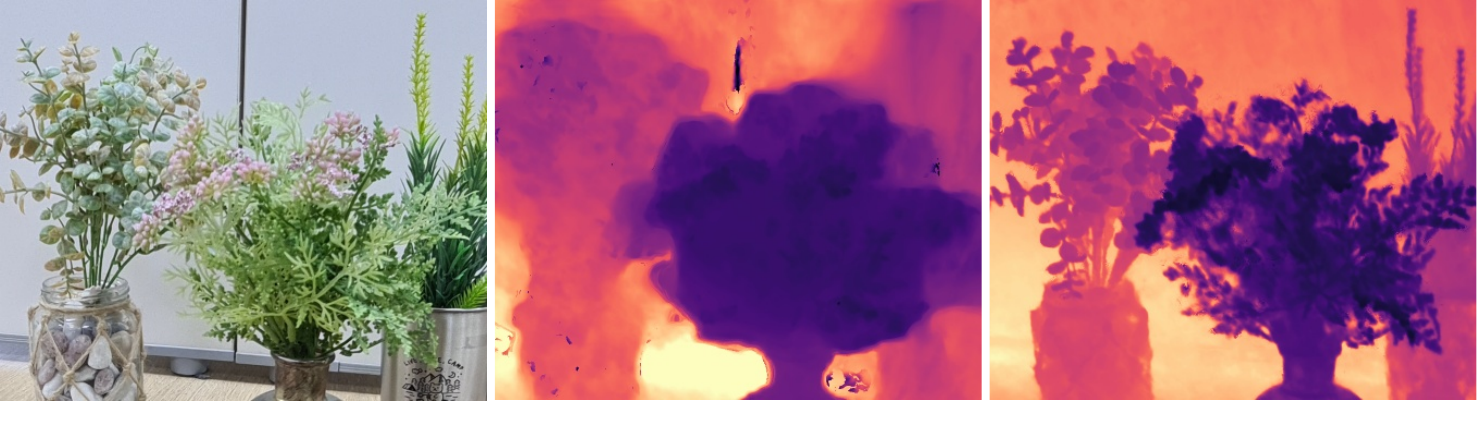}}%
    \put(0.16450636,0.00890198){\color[rgb]{0,0,0}\makebox(0,0)[t]{\smash{\begin{tabular}[t]{c}(a) RGB\end{tabular}}}}%
    \put(0.49955811,0.00890198){\color[rgb]{0,0,0}\makebox(0,0)[t]{\smash{\begin{tabular}[t]{c}(b) NerfingMVS\end{tabular}}}}%
    \put(0.83467295,0.00890198){\color[rgb]{0,0,0}\makebox(0,0)[t]{\smash{\begin{tabular}[t]{c}(c) Ours\end{tabular}}}}%
  \end{picture}%
\endgroup%
}
    \caption{\label{fig:nerfingmvs_vs_ours}
        NerfingMVS~\cite{wei2021nerfingmvs}'s results before filtering show more artifacts than ours.
    }
\end{figure}
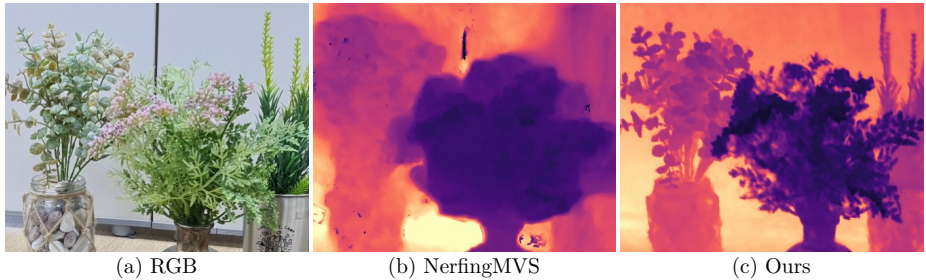

\mparagraph{Design Choices and Our Approach}
Despite improving on numerous aspects such as optimization time and novel view quality over neural representations, Plenoxels~\cite{yu2021plenoxels} produces fuzzy depth maps \NEW{(see the supplemental material).}%
We also observe that NerfingMVS's~\cite{wei2021nerfingmvs} guided NeRF optimization based on reducing the sampling range around the expected depth tends to create artifacts (Figure~\ref{fig:nerfingmvs_vs_ours}) that require an additional filtering step~\cite{Valentin2018filter} for accurate depth maps. Since filtering at novel views is impossible, this approach is not suitable for creating training data for unseen views. 
Thus, we use MipNeRF~\cite{barron2021mipnerf} as a basis for our multiview depth estimation pipeline. This can naturally handle the resolution difference between the ToF and RGB cameras.
For depth supervision~\cite{kangle2021dsnerf}, we use a straightforward approach similar to VideoNeRF~\cite{xian2021space}'s supervision on inverse depth maps computed from multi-view stereo: we add a loss using depth samples to the optimization:
\begin{equation}
    \mathcal{L}_{depth} = \omega\left\|d_{\text{rendered}} - d_{\text{ToF}}\right\|_2^2,
\end{equation}
with the ToF confidence $\omega$  (Section~\ref{sec:tof_depth}).
In addition to this supervision, we implement extrinsic and intrinsic refinement~\cite{jeong2021SCNeRF} following a 6D continuous rotation representation~\cite{zhou2019rotation}. Note that, since Mip-NeRF prevents too high frequencies in positional encoding and poses and that the intrinsics are well initialized, a coarse-to-fine positional encoding~\cite{lin2021barf} is not required.
Our training and validation datasets have eight scenes with around 100 snapshots per scene. The scenes feature varied depth range, background, objects, and materials.

\section{Results}

\subsection{Evaluation Dataset}
To evaluate our method, we build a real-world dataset with ground-truth depth obtained using a Kinect Azure. Since the depth camera is higher power, noise is reduced, and depth quality is much better than our phone's ToF module. 
After securing the phone and the Azure Kinect on a joint mount, we calibrate the phone's ToF module and the ultrawide camera w.r.t. the depth camera. Once the calibration is estimated, depth maps can be reprojected to the ultrawide and ToF cameras for comparison.
We capture four scenes for a total of 200 snapshots.

In addition to this RGB-D dataset, we calibrate the floating camera using the conventional multi-shot offline pipeline with chessboards. This provides ground truth for our online calibration evaluation on four scenes.

\subsection{Depth Estimation for Training}

\begin{table}[t]
	\centering
	\caption{\label{tab:dataset_eval}
		We compare our multiview fusion against a ToF-supervised scene representation~\cite{attal2021torf} and multiview-stereo approaches~\cite{wei2021nerfingmvs,kopf2021rcvd,luo2020cvd}. Original Mip-NeRF~\cite{barron2021mipnerf} on which our implementation is based is given for reference.
	}
		\begin{tabular}{l rr rrrr}
			\toprule
            & \multicolumn{2}{c}{Bad ratio (\%)} & \multicolumn{4}{c}{Depth error}  \\
            & $>$0.2 & $>$0.05 & MAE Rel. & RMSE Rel.  & MAE & RMSE \\
            \midrule
            TöRF~\cite{attal2021torf}  & {1.82} & {26.48} & {0.055} & {0.075} & {0.041} & {0.062} \\
            NerfingMVS~\cite{wei2021nerfingmvs} & {3.37} & {20.11} & {0.047} & {0.069} & {0.039} & {0.071} \\
            RCVD~\cite{kopf2021rcvd}  & {35.12} & {75.68} & {0.249} & {0.326} & {0.208} & {0.315} \\
            CVD~\cite{luo2020cvd}  & {1.68} & {11.67} & {0.033} & {0.056} & {0.028} & {0.056} \\
            Mip-NeRF~\cite{barron2021mipnerf} & {17.47} & {61.71} & {0.156} & {0.199} & {0.115} & {0.170}\\
            \midrule
            Ours & \textBF{0.81} & \textBF{7.07} & \textBF{0.028} & \textBF{0.047} & \textBF{0.022} & \textBF{0.044}\\
			\bottomrule
		\end{tabular}
\end{table}

\begin{figure}[t]
    \centering
    \resizebox{\columnwidth}{!}{\begingroup%
  \makeatletter%
  \providecommand\color[2][]{%
    \errmessage{(Inkscape) Color is used for the text in Inkscape, but the package 'color.sty' is not loaded}%
    \renewcommand\color[2][]{}%
  }%
  \providecommand\transparent[1]{%
    \errmessage{(Inkscape) Transparency is used (non-zero) for the text in Inkscape, but the package 'transparent.sty' is not loaded}%
    \renewcommand\transparent[1]{}%
  }%
  \providecommand\rotatebox[2]{#2}%
  \newcommand*\fsize{\dimexpr\f@size pt\relax}%
  \newcommand*\lineheight[1]{\fontsize{\fsize}{#1\fsize}\selectfont}%
  \ifx\svgwidth\undefined%
    \setlength{\unitlength}{425.19685039bp}%
    \ifx\svgscale\undefined%
      \relax%
    \else%
      \setlength{\unitlength}{\unitlength * \real{\svgscale}}%
    \fi%
  \else%
    \setlength{\unitlength}{\svgwidth}%
  \fi%
  \global\let\svgwidth\undefined%
  \global\let\svgscale\undefined%
  \makeatother%
  \begin{picture}(1,0.38)%
    \lineheight{1}%
    \setlength\tabcolsep{0pt}%
    \put(0,0){\includegraphics[width=\unitlength,page=1]{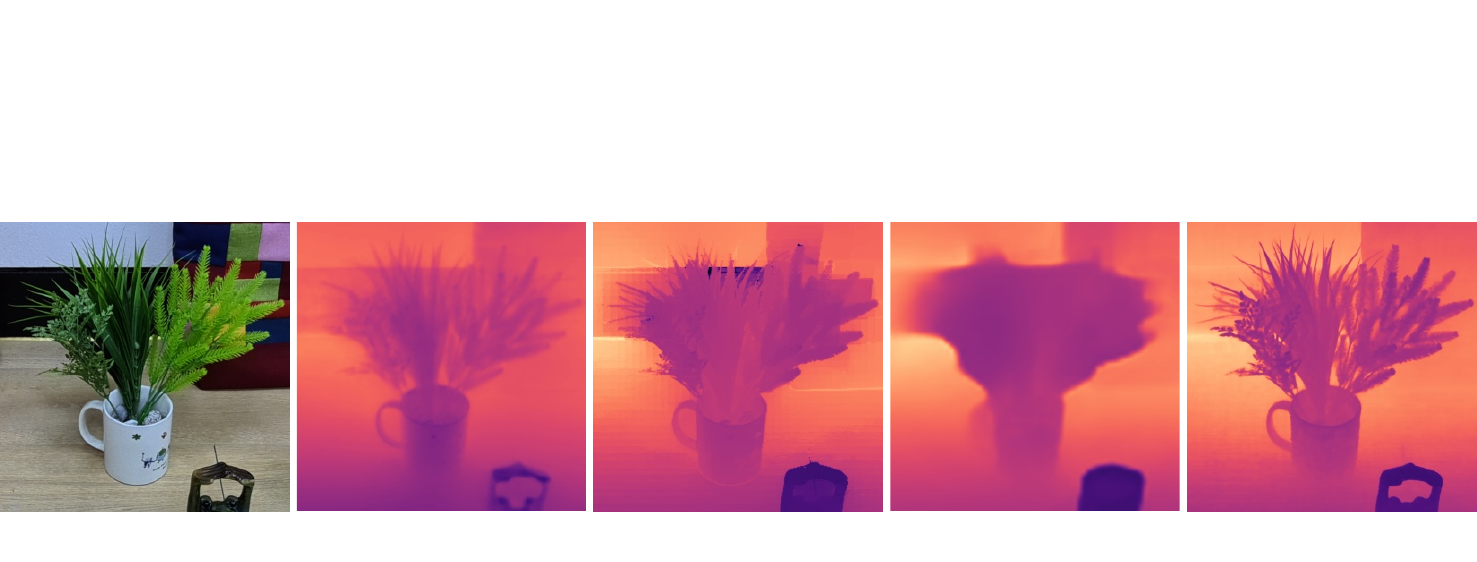}}%
    \put(0.09734985,0.00823903){\color[rgb]{0,0,0}\makebox(0,0)[t]{\smash{\begin{tabular}[t]{c}(a) RGB\end{tabular}}}}%
    \put(0.49933716,0.00823903){\color[rgb]{0,0,0}\makebox(0,0)[t]{\smash{\begin{tabular}[t]{c}(c) NerfingMVS\end{tabular}}}}%
    \put(0.90141917,0.00823903){\color[rgb]{0,0,0}\makebox(0,0)[t]{\smash{\begin{tabular}[t]{c}(e) Ours\end{tabular}}}}%
    \put(0.29885244,0.00823903){\color[rgb]{0,0,0}\makebox(0,0)[t]{\smash{\begin{tabular}[t]{c}(b) TöRF\end{tabular}}}}%
    \put(0.70050832,0.00823903){\color[rgb]{0,0,0}\makebox(0,0)[t]{\smash{\begin{tabular}[t]{c}(d) CVD\end{tabular}}}}%
    \put(0,0){\includegraphics[width=\unitlength,page=2]{mvs_comparison.pdf}}%
  \end{picture}%
\endgroup%
}
    \caption{\label{fig:mvs_comparison} 
    Multiview depth estimation with ToF-supervised scene representation TöRF~\cite{attal2021torf} and the multiview-stereo approaches NerfingMVS \cite{wei2021nerfingmvs} and CVD \cite{luo2020cvd}.
    }
	\vspace{-3mm}
\end{figure}

Figure~\ref{fig:mvs_comparison} shows that our method can preserve thin structures, and Table~\ref{tab:dataset_eval} confirms that our approach can efficiently merge ToF and stereo data from multiple views. While TöRF~\cite{attal2021torf} is designed to handle ToF inputs, it performs worse than our method when ToF and RGB resolutions differ.
We run CVD and RCVD at their default resolution since we observed degradation in accuracy when increasing image size.
Note that progressive geometry integration methods such as~\cite{Ha2021NormalFusion} fail on our data: the raw ToF depth maps from our smartphone are not accurate enough for ICP registration.

\subsection{Snapshot RGB-D Imaging}

\begin{table}[t]
	\centering
	\caption{\label{tab:snapshot_abla_calib}
		Our online calibration versus Gao et al.~\cite{gao2017selfcalibratingtof} on our real dataset. 
	}
	\begin{tabular}{l rr rr}
		\toprule
		&  \multicolumn{2}{c}{Pose error (mm)} &  \multicolumn{2}{c}{Rotation error (deg.)}  \\
		& MAE & RMSE & MAE & RMSE  \\
		\midrule
		Gao et al. \cite{gao2017selfcalibratingtof} & {5.664} & {5.997} & {1.292} & {1.431} \\
		\cite{gao2017selfcalibratingtof}  + DFM \cite{efe2021dfm} & {5.174} & {5.803} & {1.344} & {1.499} \\
		\midrule
        Ours  & \textBF{3.346} & \textBF{3.989} & \textBF{1.264} & \textBF{1.411}  \\
		\bottomrule
	\end{tabular}
\end{table}
\mparagraph{Calibration}
Table~\ref{tab:snapshot_abla_calib} details the accuracy of our calibration method. 
\NEW{Gao et al.~[15] calibrate two GoPro cameras w.r.t.~a Kinect RGB-D camera. In our comparison, we substitute the RGB-D camera by the ultrawide RGB plus the ToF module, and substitute the GoPro camera by the main camera. }
Gao et al.~\cite{gao2017selfcalibratingtof}'s calibration shows much lower accuracy than ours, even when the method is paired with a state-of-the-art feature matcher~\cite{efe2021dfm}. 
In addition, only our method is able to refine the camera matrix and distortion parameters.

\begin{figure}[thpb]
	\centering
	\resizebox{\columnwidth}{!}{\begingroup%
  \makeatletter%
  \providecommand\color[2][]{%
    \errmessage{(Inkscape) Color is used for the text in Inkscape, but the package 'color.sty' is not loaded}%
    \renewcommand\color[2][]{}%
  }%
  \providecommand\transparent[1]{%
    \errmessage{(Inkscape) Transparency is used (non-zero) for the text in Inkscape, but the package 'transparent.sty' is not loaded}%
    \renewcommand\transparent[1]{}%
  }%
  \providecommand\rotatebox[2]{#2}%
  \newcommand*\fsize{\dimexpr\f@size pt\relax}%
  \newcommand*\lineheight[1]{\fontsize{\fsize}{#1\fsize}\selectfont}%
  \ifx\svgwidth\undefined%
    \setlength{\unitlength}{425.19685039bp}%
    \ifx\svgscale\undefined%
      \relax%
    \else%
      \setlength{\unitlength}{\unitlength * \real{\svgscale}}%
    \fi%
  \else%
    \setlength{\unitlength}{\svgwidth}%
  \fi%
  \global\let\svgwidth\undefined%
  \global\let\svgscale\undefined%
  \makeatother%
  \begin{picture}(1,0.96)%
    \lineheight{1}%
    \setlength\tabcolsep{0pt}%
    \put(0.09734985,0.01013346){\color[rgb]{0,0,0}\makebox(0,0)[t]{\smash{\begin{tabular}[t]{c}(a) RGB\end{tabular}}}}%
    \put(0.50004741,0.01013346){\color[rgb]{0,0,0}\makebox(0,0)[t]{\smash{\begin{tabular}[t]{c}(c) Agresti et al.\end{tabular}}}}%
    \put(0.90141907,0.01013346){\color[rgb]{0,0,0}\makebox(0,0)[t]{\smash{\begin{tabular}[t]{c}(e) Ours\end{tabular}}}}%
    \put(0.2989945,0.01013346){\color[rgb]{0,0,0}\makebox(0,0)[t]{\smash{\begin{tabular}[t]{c}(b) Marin et al.\end{tabular}}}}%
    \put(0.70110016,0.01013346){\color[rgb]{0,0,0}\makebox(0,0)[t]{\smash{\begin{tabular}[t]{c}(d) Gao et al.\end{tabular}}}}%
    \put(0,0){\includegraphics[width=\unitlength,page=1]{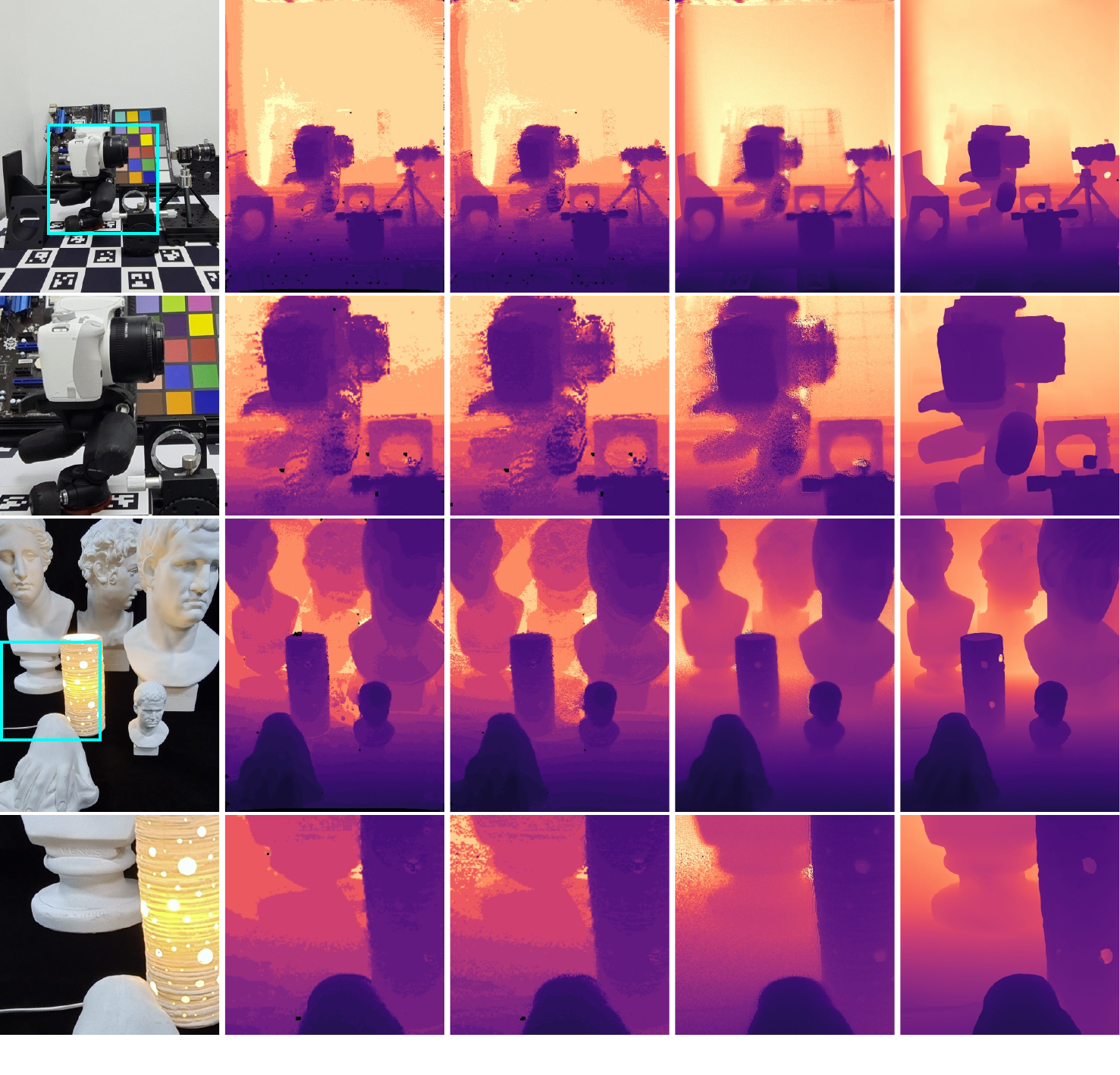}}%
  \end{picture}%
\endgroup%
}
	\caption{\label{fig:snapshot_comparison}
		ToF/stereo fusion results. We pair all other methods with a state-of-the-art stereo approach for fair comparison \cite{lipson2021raft-stereo}. Marin et al.~\cite{Marin2016ReliableFO} and Agresti et al.~\cite{agresti2017fusion} suffer from quantization as their sub-pixel resolution approach is not suitable for small phone camera baselines. Gao et al.~\cite{gao2017selfcalibratingtof} relies heavily on ToF measurements, degrading its performance when ToF is inaccurate (e.g., black parts of the camera). \NEW{Refer to the supplemental material for additional results.}
	}
\end{figure}
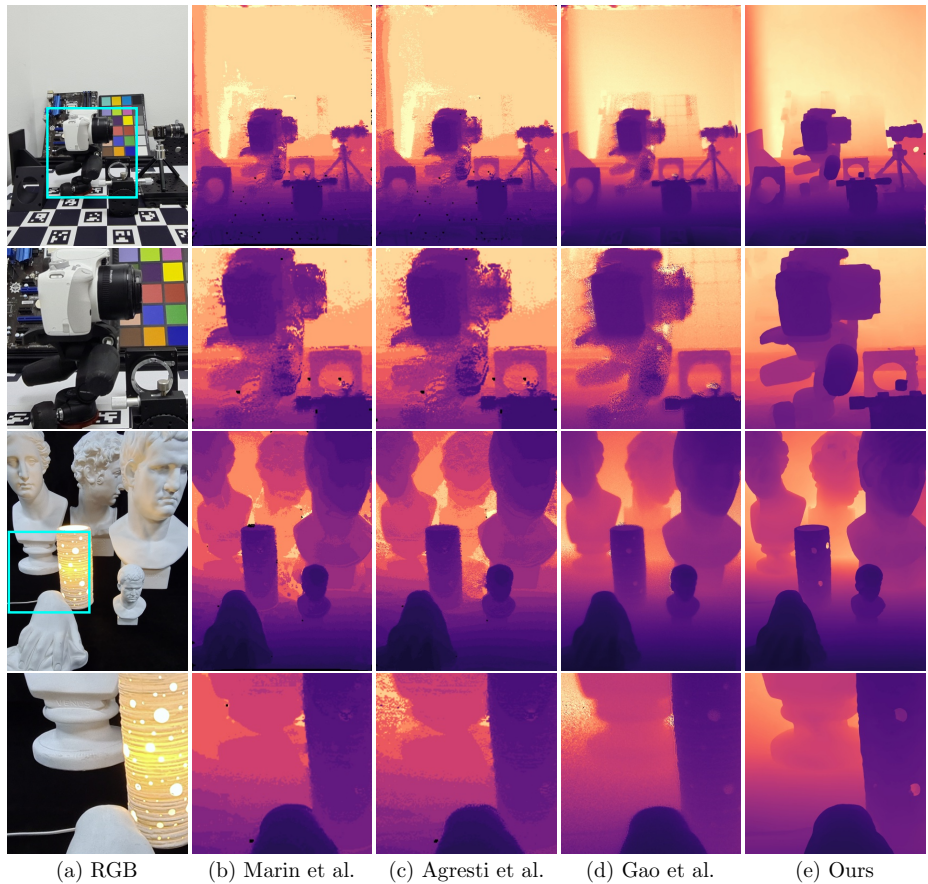

\begin{table}[t]
	\centering
	\caption{\label{tab:snapshot_comparison}
	Fusion evaluation. In the first rows, we evaluate other approaches for RGB/ToF fusions. Since their stereo matching method are not robust against noise and imaging artifacts, their fusion results are highly inaccurate.
	In the next rows, we replace their less robust stereo matching with a state-of-the-art method~\cite{lipson2021raft-stereo}. 
	We use our calibration for all methods\NEW{, except for ``Ours (ignoring OIS)'' to highlight the importance of our per-shot calibration}. 
	The results show that our fusion approach outperforms existing methods.
	}
    \begin{tabular}{l rr rrrr}
        \toprule
        &  \multicolumn{2}{c}{Bad ratio (\%)} &  \multicolumn{4}{c}{Depth error}  \\
        & $>$0.2 & $>$0.05 & MAE Rel. & RMSE Rel.  & MAE & RMSE \\
        \midrule
        Marin et al.~\cite{Marin2016ReliableFO} & {61.25} & {96.80} & {0.545} & {0.708} & {0.410} & {0.610} \\
        Agresti et al.~\cite{agresti2017fusion,agresti2019fusion_synthetic}  & {91.29} & {98.29} & {0.962} & {1.172} & {0.722} & {1.012}\\
        Gao et al.~\cite{gao2017selfcalibratingtof}  & {14.95} & {40.82} & {2.262} & {6.493} & {1.394} & {5.066}\\
        \midrule
		\cite{Marin2016ReliableFO} (RAFT-stereo~\cite{lipson2021raft-stereo}) & {2.01} & {13.87} & {0.043} & {0.092} & {0.034} & {0.085} \\
		\cite{agresti2017fusion,agresti2019fusion_synthetic} (RAFT-stereo~\cite{lipson2021raft-stereo}) & {1.48} & {13.96} & {0.037} & {0.065} & {0.031} & {0.065} \\
		\cite{gao2017selfcalibratingtof} (RAFT-stereo~\cite{lipson2021raft-stereo}) & {1.67} & {7.71} & {0.031} & {1.021} & {0.026} & {0.854} \\
        \midrule
		\NEW{Stereo only~\cite{lipson2021raft-stereo}} & {2.36} & {14.18} & {0.041} & {1.300} & {0.035} & {1.078} \\
		\NEW{Ours (ignoring OIS)}  & {9.06} & {29.54} & {0.082} & {0.163} & {0.073} & {0.164} \\
		\midrule
        Ours & \textBF{1.40} & \textBF{7.17} & \textBF{0.028} & \textBF{0.050} & \textBF{0.024} & \textBF{0.051}\\
        \bottomrule
    \end{tabular}
    \vspace{-0.35cm}
\end{table}

\mparagraph{Stereo/ToF Fusion}
We evaluate our fusion approach against our real-world RGB-D dataset. For comparison, we implement~\cite{Marin2016ReliableFO,agresti2017fusion,gao2017selfcalibratingtof} and we train Agresti et al.'s method~\cite{agresti2017fusion} using their rendered SYNTH3 dataset. 
Since Gao et al.~\cite{gao2017selfcalibratingtof}'s calibration is too inaccurate, rectification fails severely on some snapshots. 
\NEW{
Total calibration failure occurred for 34 of the 200 snapshots (17\%) in our test dataset. We show examples of poorly rectified stereo pairs in the supplemental material.
}

Thus, evaluate all methods using our calibration.
\NEW{
We also evaluate if we can ignore OIS: we calibrate the main camera using a checkerboard while the phone is fixed, then we move the phone to capture our test scenes. We report the results in Table~\ref{tab:snapshot_comparison} under ``Ours (ignoring OIS)'', showing a large decrease in depth accuracy. Thus, online calibration is both necessary and effective.
}
Figure~\ref{fig:snapshot_comparison} shows that our method allows for robust depth estimation with better edge and hole preservation. The low RMSE in Table~\ref{tab:snapshot_comparison} suggests that our method is robust against strong outliers. While other methods suffer from a less robust stereo matching\NEW{---swapping theirs for RAFT-stereo~\cite{lipson2021raft-stereo} significantly improving their results---}our approach maintains higher accuracy.

\NEW{\subsection{Dependency on the Device}
Under the assumption of a narrow baseline on smartphones, the method should generalize as it allows accurate optical flow estimation between the two RGB cameras. In addition, we show the results of our fusion of datasets based on different hardware:
a ZED stereo camera and a Microsoft Kinect v2 ToF depth camera for REAL3~\cite{agresti2019fusion_synthetic}, and two calibrated BASLER scA1000 RGB cameras and
a MESA SR4000 ToF camera for LTTM5 \cite{mutto2015probabilisticFusion} in the supplemental material.}

\subsection{Limitations}
\label{sec:limitations}

While our approach applies to indoor environments, the reliance on ToF and stereo prevents application in some scenarios. First, the ToF module cannot estimate depth accurately at large distances due to its low power. Second, ToF depth estimation is not reliable within strong IR ambient illumination (e.g., direct daylight). Since our calibration relies directly on ToF measurements, it becomes inaccurate if no ToF depth can be estimated. In addition, some materials---particularly translucent or specular materials---are challenging for both ToF and stereo depth estimation and cannot be tackled by our fusion approach.

\section{Conclusion}

Optical-image-stabilized lenses are now common but present problems for pose estimation when wanting to fuse information across multiple sensors in a camera system. 
This limits our ability to estimate high-quality depth maps from a single snapshot. 
Our method is designed for consumer devices, tackling calibration and robust sensor fusion for indoor environments.
As our approach uses only a single snapshot and does not exploit camera motion for pose estimation, the acquisition is quick and could be used on dynamic scenes.
\NEW{Evaluated on real-world inputs, our method estimates more accurate depth maps than state-of-the-art ToF and stereo fusion methods.}

\section*{Acknowledgement}
Min H.~Kim acknowledges funding from Samsung Electronics, in addition to the partial support of the MSIT/IITP of Korea (RS-2022-00155620, 2022-0-00058, and 2017-0-00072), the NIRCH of Korea (2021A02P02-001),  Microsoft Research Asia, and the Samsung Research Funding Center (SRFC-IT2001-04) for developing 3D imaging algorithms. James Tompkin thanks US NSF CAREER-2144956.

\clearpage
\bibliographystyle{splncs04}
\bibliography{bibliography}

\end{document}